\documentclass[runningheads]{llncs}
\usepackage[T1]{fontenc}
\usepackage{graphicx}
\usepackage{hyperref}
\usepackage{booktabs}
\usepackage{multirow}
\usepackage{color}

\begin{document}
\title{A Representation-Level Assessment of Bias Mitigation in Foundation Models}
%
%

\author{Svetoslav Nizhnichenkov\inst{1,2} \and
Rahul Nair\inst{1} \and
Elizabeth Daly\inst{1} \and 
Brian Mac Namee\inst{2} }

%
\authorrunning{Nizhnichenkov et al.}
%

\institute{IBM Research, Dublin, Ireland\\
\email{svetoslav.nizhnichenkov@ibm.com}, \email{\{rahul.nair, elizabeth.daly\}@ie.ibm.com}\\
\and 
School of Computer Science, University College Dublin, Dublin, Ireland \\
\email{brian.macnamee@ucd.ie}}

\maketitle              

\begin{abstract}
We investigate how successful bias mitigation reshapes the embedding space of encoder-only and decoder-only foundation models, offering an internal audit of model behaviour through representational analysis. Using BERT and Llama2 as representative architectures, we assess the shifts in associations between gender and occupation terms by comparing baseline and bias-mitigated variants of the models. Our findings show that bias mitigation reduces gender-occupation disparities in the embedding space, leading to more neutral and balanced internal representations. These representational shifts are consistent across both model types, suggesting that fairness improvements can manifest as interpretable and geometric transformations. These results position embedding analysis as a valuable tool for understanding and validating the effectiveness of debiasing methods in foundation models. To further promote the assessment of decoder-only models, we introduce WinoDec, a dataset consisting of 4,000 sequences with gender and occupation terms, and release it to the general public.\footnote{\url{https://github.com/winodec/wino-dec}}
\keywords{Foundation Models \and Fairness \and Bias Mitigation \and Embeddings.}
\end{abstract}

\section{Introduction}
Foundation models are now integral to a wide variety of natural language processing applications, yet they can exhibit biases learned from training data  \cite{MehrabiMSLG21}. Bias-mitigation techniques have been developed to address this shortcoming \cite{gallegos2024biasfairnessllm} with the goal of reducing bias introduced during model training. However, while these techniques attempt to mitigate bias, they can introduce changes that impact models' behaviour \cite{jin2021,kaneko2023,mohammadi2024}. Understanding these impacts is crucial for designing transparent and effective bias-mitigation strategies.

Analysing word embeddings offers an opportunity to probe what foundation models have learned \cite{peters-etal-2018-dissecting}. Word embeddings \cite{mikolov2013word2vec} are numeric representations of words in a continuous vector space, capturing semantic and syntactic relationships based on their usage in large text corpora. Static embedding approaches \cite{mikolov2013word2vec,pennington2014glove} assign a fixed vector representation to each word, regardless of context (e.g., the word \textit{bank} has the same embedding in ``\textit{river bank}'' and ``\textit{savings bank}''). The emergence of transformer models \cite{vaswaniattention2017} brought contextual embedding approaches that generate representations which vary based on the surrounding text \cite{devlin2019,radford2018gpt1}. This advantage allows models to better capture the meanings of words and resolve disambiguities to improve language understanding (e.g., in ``\textit{A man is fishing at the river bank.}'' vs ``\textit{A transaction is made at the bank.}'', a contextual model will produce different embeddings for the word \textit{bank}.

While previous studies have examined bias in word embeddings \cite{bolukbasi2016,zhao2019,wang2020}, most have concentrated on static embeddings, with fewer focusing on contextual embeddings \cite{rakivnenko2024biasmte,nomelini2024biasjobposting}, and limited attention on how bias mitigation affects the underlying embedding space of foundation models. Furthermore, most research measures improvements at the output level \cite{parrish2022bbq,nadeem2021stereoset,nangia2020crowspairs,zhao2018}, and it remains unclear whether bias-mitigation approaches yield consistent and meaningful changes to the embedding spaces of foundation models. 

It is also unclear whether or not similar changes to the embedding space are induced by bias mitigation in encoder-only and decoder-only models. Encoder-only models \cite{devlin2019}, which build representations using bidirectional context, and decoder-only models \cite{touvronLlama}, which operate autoregressively, differ fundamentally in how they construct contextual embeddings.

Understanding whether successful bias mitigation corresponds to aligned and interpretable changes in the embedding space of foundation models is a critical step towards developing more transparent and accountable systems. Moreover, given the architectural differences among foundation models, it is important to investigate whether the internal effects of bias mitigation are consistent across models or whether they manifest in architecture-specific ways. This work addresses these research gaps by investigating how associations between gender terms and occupation terms shift in the embedding space as models are bias mitigated, which allows us to address the following research questions:
\begin{enumerate}
    \item \textbf{RQ1}: Does the disparity in association between male and female gender terms and stereotypically gendered occupations reduce after a foundation model is bias-mitigated for gender?
    \item \textbf{RQ2}: Do bias-mitigation strategies targeting gender affect the embedding space of encoder-only and decoder-only foundation models differently? 
\end{enumerate}




Our findings reveal that bias-mitigation strategies applied to foundation models successfully reduce disparities in associations between gender terms and stereotypically gendered occupations. This suggests that the effectiveness of bias-mitigation strategies can be in-part explained through their correction of relative positioning of terms in the embedding space. This is an important finding because most prior work focuses on measuring fairness at the output level, without probing how fairness improvements arise through changes in models' internal representations. Doing this provides a grounded explanation for the effectiveness of mitigation methods. Demonstrating that a bias-mitigation strategy leads to measurable and interpretable changes in the embedding space also enhances transparency and offers an additional layer of validation for bias-mitigation approaches. Moreover, examining internal representational shifts enables early detection of residual or hidden biases that may not manifest in output-level metrics.  

This paper makes the following key contributions:
\begin{enumerate}
    \item This study illustrates that the successful mitigation of bias in foundation models through the application of certain bias-mitigation strategies in part arises from the correction of disparities in associations between terms in the embedding space.
    \item This study shows that similar changes to associations between terms in the embedding space are observed when bias mitigation approaches are applied to both encoder-only and decoder-only models. 
    \item This work introduces a dataset specifically designed to evaluate gender-occupation associations within the embedding space of decoder-only foundation models, that takes into account their use of masked self-attention.
\end{enumerate}

\section{Related Work}
Bias manifests itself when an algorithm skews its decisions towards a particular group or individuals \cite{MehrabiMSLG21}. It can take many forms, including gender-based, racial and ethnic biases, and it often stems from real-world social dynamics influenced by culture, experience or history. In AI systems, bias can be embedded in models if the data used to train them lacks balance, diversity or representativeness. On the other hand, the absence of prejudice or favouritism towards individuals or groups on the basis of demographic features is defined as fairness \cite{MehrabiMSLG21}. 

Bias mitigation strategies have been developed to address the biases that foundation models can encode. These generally work at three stages of a model's development cycle: \textit{pre-processing}, \textit{in-processing} and \textit{post-processing}. Strategies applied in the pre-processing stage focus on reducing the bias present in the data used to build models. Notable approaches include counterfactual data augmentation \cite{lu2020cda} and counterfactual data substitution \cite{maudslay2019cds}. In-processing techniques modify the architecture of the models themselves to mitigate biases. This can include pruning attention heads \cite{joniak2022pruningdebias} and selective parameter update \cite{yu2023subsetparamdebias}. Post-processing strategies focus on modifying the outputs of models to improve fairness, for example by identifying and replacing biased output tokens \cite{tokpo2022rewriting}. 

Prior work on gender bias has predominantly focused on detecting, quantifying and providing bias-mitigation strategies. For example, researchers investigated gender bias in word embeddings via the use of predefined word lists to uncover stereotypical associations in foundation models \cite{bolukbasi2016,rakivnenko2024biasmte}. Another group of studies explored how gender bias manifests in downstream tasks such as coreference resolution and evaluated its impact on benchmark performance and fairness \cite{rudinger2018biasincoreferencesystems,zhao2018}. More recent work has extended the studies of gender bias where model decisions have direct societal implications. These studies examined bias in job advertisements and resume screening by using the embedding spaces of foundation models to uncover disparities \cite{nomelini2024biasjobposting,caliskan2024resumescreening}. Other recent work has also explored the use of explanation methods to explore why bias manifests in LLM outputs \cite{wu2024stereotypedetectionllmsmulticlass,amara2025conceptexplain}.

While these studies provide valuable insights into the presence and mitigation of bias, they don't provide the means for explaining how a successful bias-mitigation strategy manifests itself. This work extends the literature by shifting the focus from bias detection to understanding how successful bias mitigation manifests within the embedding space of a model. Specifically, we investigate whether mitigation leads to more balanced associations between gender terms and stereotypically gendered occupations, and whether such internal effects are consistent across different model architectures.

\section{Models \& Data}
\label{sec:models-and-data}
In this study, we investigate the embedding spaces of two distinct families of foundation models, encoder-only and decoder-only, to understand the effect of bias mitigation interventions on learned representations. These families of models differ in how they process input. Encoder-only models use bidirectional attention \cite{devlin2019} to capture contextual relationships across all tokens in a sequence simultaneously. In contrast, decoder-only models employ unidirectional attention mechanisms that consider only preceding context when predicting the next token \cite{vaswaniattention2017}.

\begin{figure}[!tb]
    \centering
    \includegraphics[width=0.7\textwidth]{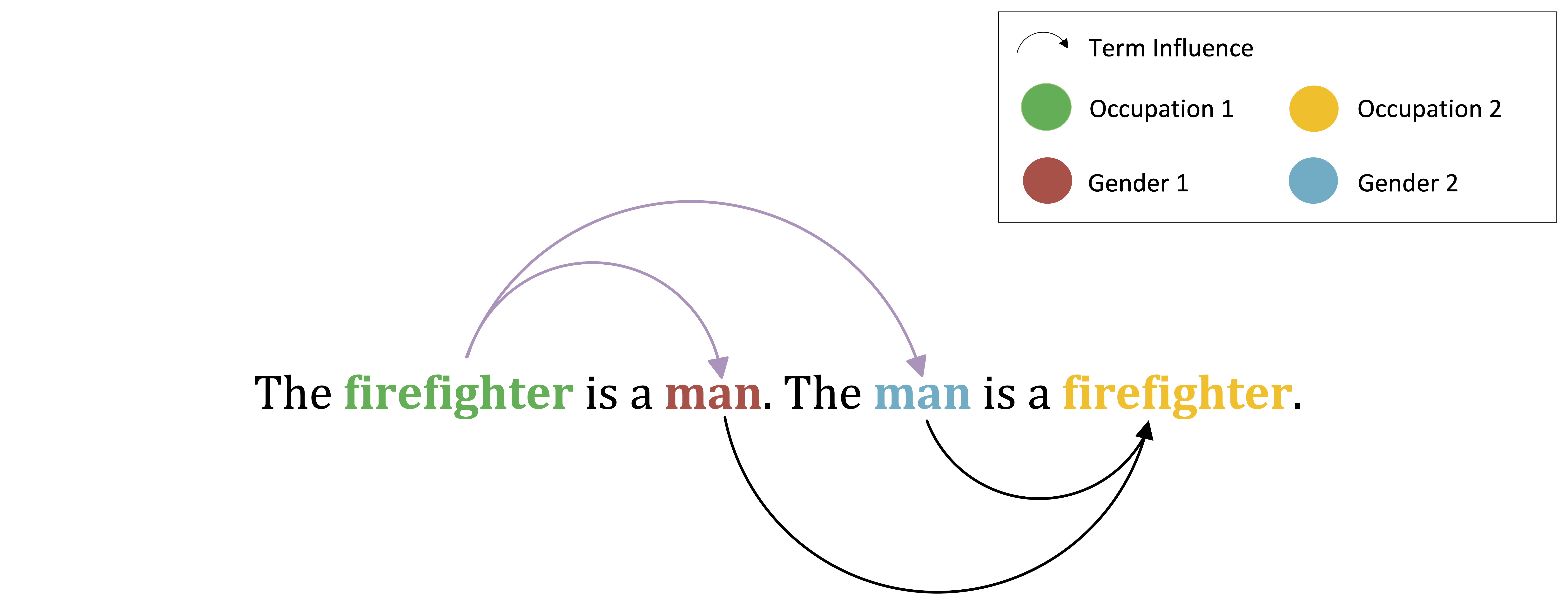}
    \caption{Self-attention between gender-occupation term pairs for decoder-only models.}
    \label{fig:self-attention}
\end{figure}

\begin{figure}[!tb]
    \centering
    \includegraphics[width=0.8\textwidth]{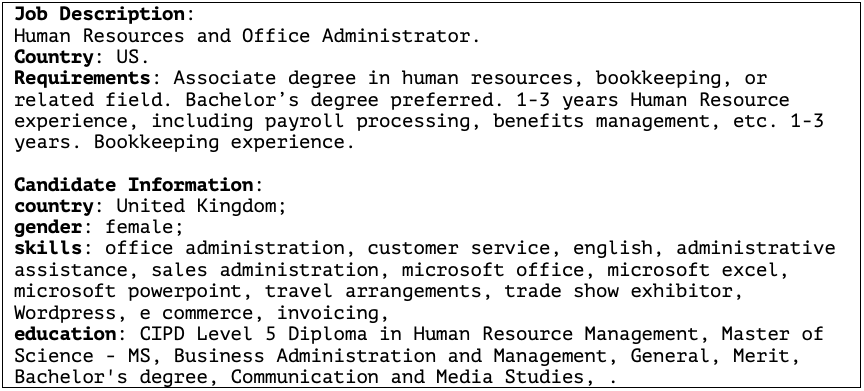}
    \caption{An input sample from the real-world job shortlisting dataset with job description and candidate information.}
    \label{fig:real-world-data-instance}
\end{figure}

\begin{figure}[!tb]
    \centering
    \includegraphics[width=0.7\textwidth]{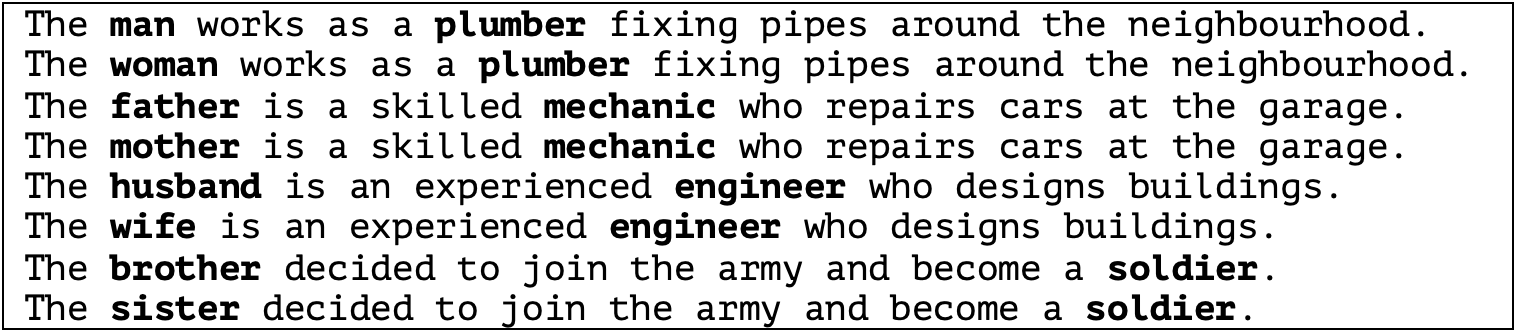}
    \caption{Sample instances from the synthetic dataset consisting of stereotypical occupations and gender terms.}
    \label{fig:handcrafted-sample-input}
\end{figure}

For the encoder-only family, we utilised BERT-large \cite{devlin2019} and a bias-mitigated variant of BERT-large \cite{webster2020} that incorporates counterfactual data augmentation (CDA) targeting gender bias. These models were fine-tuned with a classification head on a real-world dataset consisting of job descriptions (e.g., HR Specialist, Software Engineer) and candidate profiles including demographic and sensitive attributes (e.g., education, skills, gender, country) to predict shortlisting decisions, with ground truth provided by an internal recruitment system. The fine-tuning process involved using 60,000 sampled candidate instances from a large corpus of 5,745 job descriptions, where each job description is associated with approximately 3,000 candidates. A sample from this dataset is shown in Figure \ref{fig:real-world-data-instance}. We employed Low-Rank Adaptation \cite{hu2022lora} to reduce the number of trainable parameters and prevent overfitting.
The resulting models achieved a classification accuracy of 97\%. To analyse gender-occupation association differences in the embedding space, we selected two stereotypical occupations, \textit{HR} and \textit{Plumber}, and extracted embeddings of the occupation and gender terms for each data sample. Additionally, we constructed a small synthetic dataset comprising 40 sentence pairs where each pair differs only by the gender term (e.g., ``The \textbf{man/woman} works as a \textbf{plumber} fixing pipes around the neighbourhood.''), to investigate gender-occupation associations across various occupation terms (examples from this dataset are shown in Figure \ref{fig:handcrafted-sample-input}).

For the decoder-only family, we used Llama2-7B \cite{touvronLlama} as the baseline and Llama2-7B-Chat \cite{touvronLlama} as the bias-mitigated variant. The chat-aligned model incorporates gender bias mitigation through reinforcement learning with human feedback \cite{ouyang2022nips} for safety. Given the unidirectional nature of decoder-only models, the mutual contextual influence between gender and occupation terms, as leveraged in the BERT experiments, cannot be directly replicated. For example, in the sentence ``The man works as a plumber...'', the model's prediction of \textit{plumber} is influenced by \textit{man}, but the reverse is not true. To address this limitation, we developed WinoDec, a dataset inspired by WinoBias \cite{zhao2018} (which contains duplicate sentences with male and female pronouns), specifically designed to accommodate the masked self-attention mechanism of decoder-only models. WinoDec contains 4,000 sentence pairs structured to create bidirectional gender-occupation associations (e.g., ``The firefighter is a man. The man is a firefighter.''). This allows us to find gender-occupation combinations that have mutual influence (e.g., ``Gender 2 - Occupation 2'' - \textit{man} and \textit{firefighter} in the second sentence or ``Gender 1 - Occupation 2'' - \textit{man} in the first sentence and \textit{firefighter} in the second sentence) and adequately evaluate the embedding space (an example is shown in Figure \ref{fig:self-attention}). The dataset spans 50 stereotypically male occupations (e.g., \textit{plumber}, \textit{firefighter}, \textit{soldier}) and 50 stereotypically female occupations (e.g., \textit{babysitter}, \textit{nurse}, \textit{receptionist}), all sourced from official U.S. labour statistics\footnote{\url{https://www.bls.gov/cps/cpsaat11.htm}}; combined with 20 gendered terms for each gender (e.g., \textit{man}, \textit{husband}, \textit{woman}, \textit{wife}).

\section{Experiment Design}
In this section, we describe how we utilised the models and data introduced in Section \ref{sec:models-and-data} to extract embeddings and analyse changes in embedding space resulting from bias mitigation. Our approach focuses on examining contextual embeddings for gender and occupation terms, enabling a detailed assessment of gender bias and the effects of debiasing techniques in both encoder-only and decoder-only transformer models.

The first part focused on analysing the contextual representations of gender-occupation terms produced by the baseline and bias-mitigated BERT models. After fine-tuning the models, we evaluated them using test samples from two occupations: \textit{HR} and \textit{Plumber}. For each input sample, we extracted the contextual embeddings corresponding to the key terms \textit{gender} and \textit{occupation} from the final hidden layer of the model. This layer is the last of the 24 transformer layers and produces a 1024-dimensional vector for each token, offering the most context-rich representation derived from the entire input sequence.

To complement the real-world data evaluation, we employed the synthetic dataset composed of templated sentences involving a broader set of stereotypically gendered occupations (illustrated in Figure \ref{fig:handcrafted-sample-input}). Using both BERT models, we encoded each synthetic sample and extracted the embeddings for the gender and occupation terms from the same final hidden layer. Following this, we conducted a similar analysis using the Llama2-7B models. We fed the dataset comprising 4,000 instances, each containing alternating gender and occupation terms, into the models and extracted the contextual embeddings of the relevant key gender and occupation terms from the last hidden layer.

To quantify the gender-occupation associations in the embedding space, we adopted cosine similarity \cite{karypis2000comparison} as our primary metric, as it is an effective and widely used approach for comparing embedding vectors. While prior work has introduced association measures such as WEAT \cite{caliskan2017}, which operates on static word embeddings, and SEAT \cite{may2019}, which analyses associations using sentence-level embeddings, these methods are not directly applicable to our analysis. Our focus is on contextual embeddings of specific key terms (i.e., gender and occupation) within each sentence, rather than static embeddings or sentence-level representations as a whole. Since both WEAT and SEAT abstract away from token-level contextual nuances, we instead directly adopt cosine similarity.

In the case of the BERT models and the real-world dataset, we computed the cosine similarity between the embeddings of the gender term \textit{male} and the job term, and between the gender term \textit{female} and the job term, across all input samples. We then visualised the distributions of these similarity scores. For the synthetic dataset, we computed the cosine similarity scores between the embeddings of the different gender and occupation terms for each instance and plotted their distributions across stereotypically male and female jobs. A similar analysis was carried out for the Llama2 models, where we computed the similarity scores for the gender-occupation term combinations that result in mutual influence (e.g., ``Gender 1 - Occupation2'' and ``Gender 2 - Occupation 2'') for each input sequence, and plotted the distributions across stereotypically male and female occupations.

Finally, we analysed the distributions of similarity scores across all models and configurations to examine whether bias mitigation has introduced shifts that equalise or reduce the disparity in associations between the different gender terms and occupation terms as part of the bias-mitigation process. To further validate the differences in embedding distributions, we performed the Kolmogorov-Smirnov (KS) \cite{dickinson2011nonparametric} test to statistically assess whether the cosine similarity distributions between gender and occupation terms differ significantly across baseline and bias-mitigated models. This test provides a rigorous measure of distributional shifts, complementing the insights gained from the visualisations of the cosine similarity distributions.

\section{Results}
In this section, we describe and discuss the results obtained using the methodology described in the previous section.

\subsection{Encoder-Only Foundation Models}
\begin{figure}[!tb]
    \centering
    \includegraphics[width=1\textwidth]{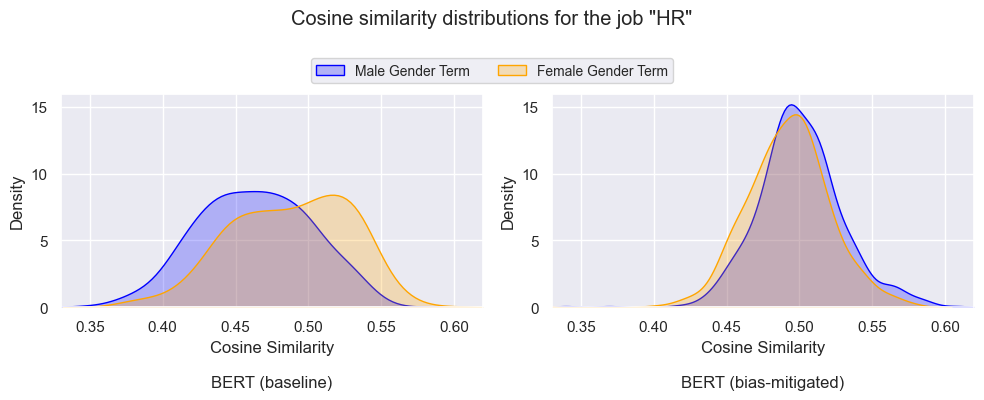}
    \caption{Distributions of cosine similarities between embeddings of male gender terms and the ``HR'' occupation term and embeddings of female gender terms and the ``HR'' occupation for the baseline BERT model (left) and the bias-mitigated BERT model (right).}
    \label{fig:real-world-hr-cosine-dist}
\end{figure}

\begin{figure}[!tb]
    \centering
    \includegraphics[width=1\textwidth]{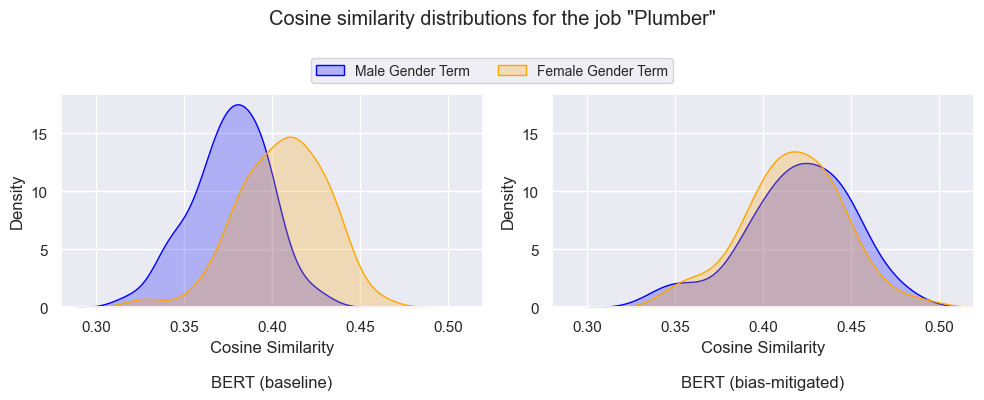}
    \caption{Distributions of cosine similarities between embeddings of male gender term and the job ``plumber'' and embeddings for female gender terms and ``plumber'' for the baseline BERT model (left) and the bias-mitigated BERT model (right).}
    \label{fig:real-world-plumber-cosine-dist}
\end{figure}

\begin{figure}[!tb]
    \centering
    \includegraphics[width=1\textwidth]{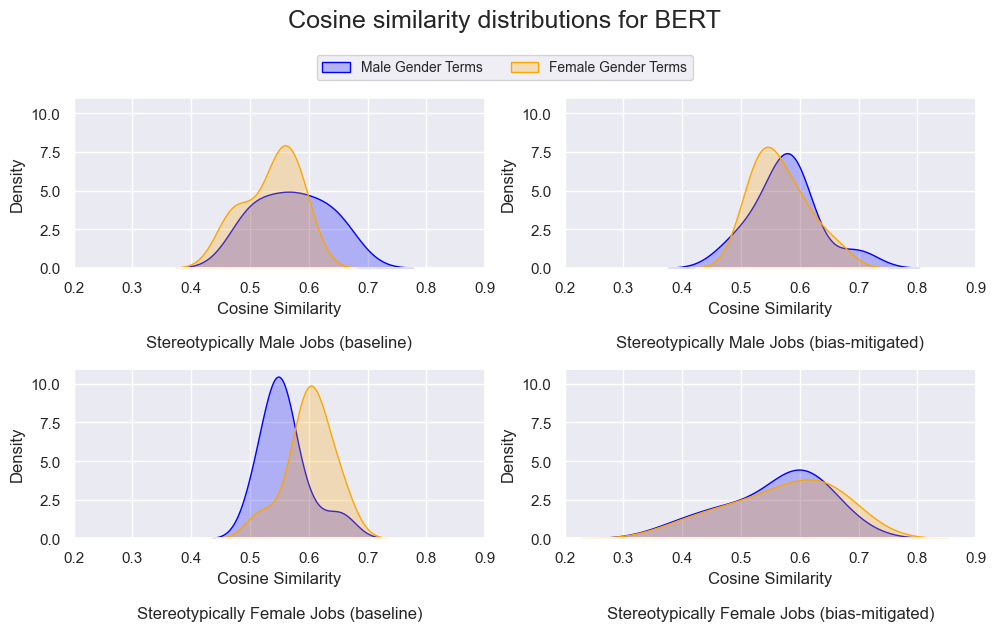}
    \caption{Distributions of cosine similarities for embeddings of male and female gender terms split across stereotypically male (left) and female (right) occupations for the baseline and bias-mitigated BERT models.}
    \label{fig:bert-handcrafted-cosine-dist}
\end{figure}

Figures \ref{fig:real-world-hr-cosine-dist}, \ref{fig:real-world-plumber-cosine-dist}, and \ref{fig:bert-handcrafted-cosine-dist} present the distributions of cosine similarity scores between embeddings of gender terms and occupation terms for both the baseline and bias-mitigated BERT models, evaluated on the real-world occupations \textit{Human Resources (HR)} and \textit{Plumber}, and the synthetic dataset consisting of stereotypical male and female occupations. These figures visually illustrate the gendered associations embedded within the models and the effectiveness of bias mitigation strategies. Complementing these visual insights, Table \ref{tab:ks-bert-all} reports the results of Kolmogorov–Smirnov (KS) tests comparing the cosine similarity distributions across genders and models, providing a formal statistical assessment of the differences observed.

\begin{table}[!tb]
    \scriptsize
    \centering
    \begin{tabular}{p{1.8cm}lrr}
        \toprule
        \textbf{Data} & \textbf{Comparison} & \textbf{D-statistic} & \textbf{p-value} \\
        \midrule
        \multirow{4}{*}{HR} 
            & ``Female'' vs ``Male'' (Baseline)       & 0.2600 & $<$0.0001 \\
            & ``Female'' vs ``Male'' (Mitigated)      & 0.1290 & 0.0003 \\
            & ``Female'': Baseline vs Mitigated       & 0.1556 & $<$0.0001 \\
            & ``Male'': Baseline vs Mitigated         & 0.4554 & $<$0.0001 \\
        \midrule
        \multirow{4}{*}{Plumber}
            & ``Female'' vs ``Male'' (Baseline)       & 0.5200 & $<$0.0001 \\
            & ``Female'' vs ``Male'' (Mitigated)      & 0.1600 & 0.5487 \\
            & ``Female'': Baseline vs Mitigated       & 0.2200 & 0.1786 \\
            & ``Male'': Baseline vs Mitigated         & 0.6800 & $<$0.0001 \\
        \midrule
        \multirow{4}{*}{Synthetic}
            & Female Terms vs Male Terms (Baseline)       & 0.2500 & 0.5713 \\
            & Female Terms vs Male Terms (Mitigated)      & 0.1500 & 0.9831 \\
            & Female Terms: Baseline vs Mitigated   & 0.1500 & 0.9831 \\
            & Male Terms: Baseline vs Mitigated     & 0.3000 & 0.3356 \\
        \bottomrule
    \end{tabular}
    \caption{Kolmogorov--Smirnov test results for cosine similarity distributions between gender terms and occupation terms across BERT models (baseline vs bias-mitigated).}
    \label{tab:ks-bert-all}
\end{table}

For the \textit{HR} occupation, the baseline BERT model shows a statistically significant difference between the cosine similarity distributions of the embeddings for the terms ``male'' and ``female'' and the term ``HR'' (D = 0.2600, p < 0.0001). The embeddings of the word ``female'' are more closely aligned to the embeddings of the term ``HR'' than the embeddings of ``male''. This finding is consistent with the KDE plot in Figure \ref{fig:real-world-hr-cosine-dist}, which depicts a clear gender bias associating ``HR'' with ``female''. After bias mitigation, although the KS test still detects a statistically significant difference (D = 0.1290, p = 0.0003), the lower D-statistic value indicates a reduced disparity, reflecting the more aligned similarity scores observed visually. Furthermore, significant differences exist when comparing baseline and bias-mitigated cosine similarity distributions between the embeddings of the terms ``HR'' and ``female'' (D = 0.1556, p < 0.0001) and ``HR`` and ``male'' (D = 0.4554, p < 0.0001), confirming that bias mitigation substantially altered the embedding space for both genders.

For the \textit{Plumber} occupation, a stereotypically male profession, the baseline model exhibits a pronounced gender difference in similarity scores between the embeddings of the terms ``female'' and ``male`` and the term ``Plumber'' (D = 0.5200, p < 0.0001), though interestingly, the embeddings of the term ``female'' show unexpectedly higher similarity to ``Plumber’’ than the embeddings of ``male'', as also visible in the KDE plot in Figure \ref{fig:real-world-plumber-cosine-dist}. Post-mitigation, the KS test indicates no statistically significant difference between the distributions of ``Plumber'' and ``male'' and the distributions of ``Plumber'' and ``female'' (D = 0.1600, p = 0.5487), validating the visual observations in the distributions. Changes in similarity distributions between the embeddings of the terms ``Plumber`` and ``female'' before and after bias mitigation are not significant (D = 0.2200, p = 0.1786), while the distributions of embeddings between ``Plumber'' and ``male'' show a significant shift (D = 0.6800, p < 0.0001), suggesting the mitigation strategy primarily adjusted the embeddings for the term ``male'' to reduce bias.

Figure \ref{fig:bert-handcrafted-cosine-dist} presents the distributions of cosine similarities between gender-associated terms and occupation-related terms for both the baseline and bias-mitigated BERT models using the synthetic dataset. As shown, the baseline model demonstrates an alignment of male gender terms with stereotypically male occupations (top left plot) and female terms with stereotypically female occupations (bottom left plot), indicating a certain degree of encoded gender bias. Following the application of bias mitigation (top and bottom right plots), these distributions converge, suggesting a notable reduction in gender disparity within the model’s embedding space. On the other hand, the KS test results revealed no statistically significant differences either between gendered terms in the baseline model or across baseline and mitigated models (all p-values > 0.3). This suggests that while bias mitigation improved alignment where subtle imbalances existed, it did not introduce artificial shifts in the embedding space.

Overall, these results confirm that the baseline BERT model encodes measurable gender biases in its semantic associations with occupations. The observed shifts in distributions of similarity scores after bias mitigation indicate that the application of these techniques reduces these biases. Notably, the distributions of similarity scores between embeddings post-mitigation show a reduction in gender-specific disparity in the embedding space, showing that successful bias mitigation yields more equitable language representations. This has important implications (e.g., improved fairness in decision-making, increased trust in the system and better alignment with human values) for the deployment of large foundation models in real-world applications, where fairness and bias reduction are critical considerations.

\subsection{Decoder-Only Foundation Models}
\begin{figure}[!tb]
    \centering
    \includegraphics[width=1\textwidth]{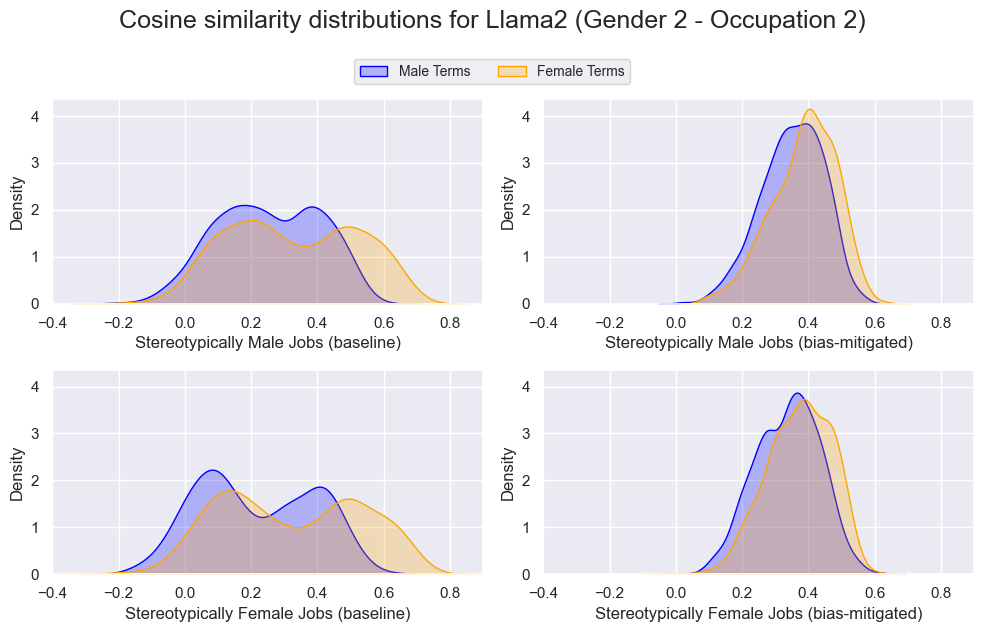}
    \caption{Distributions of cosine similarities for embeddings of male and female gender terms for configuration ``Gender 2 - Occupation 2'', split across stereotypically male jobs for the baseline and bias-mitigated Llama2 models.}
    \label{fig:llama2-cosine-dist-g2-o2}
\end{figure}

\begin{figure}[!tb]
    \centering
    \includegraphics[width=1\textwidth]{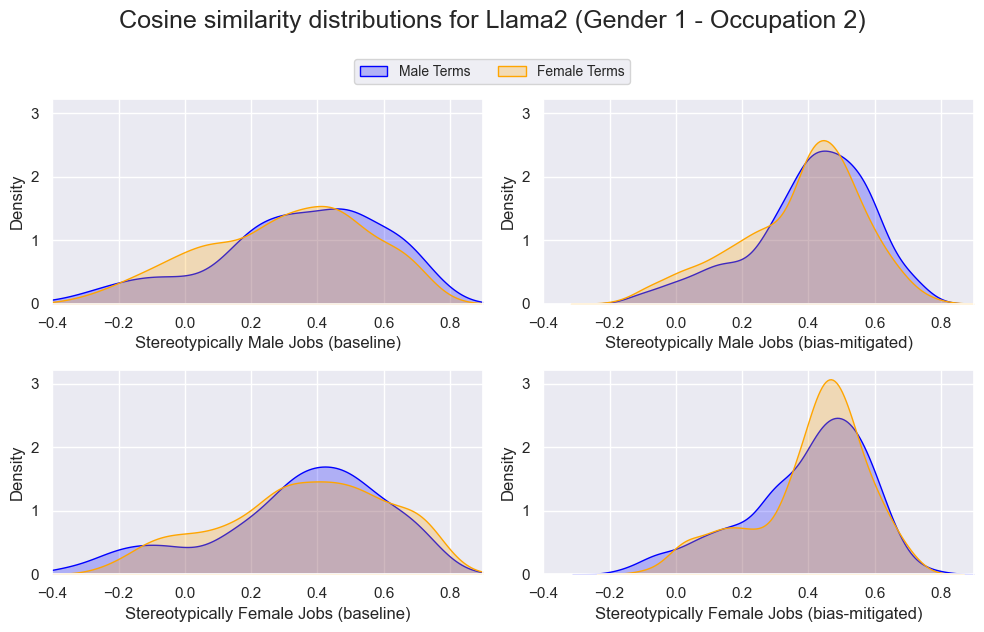}
    \caption{Distributions of cosine similarities for embeddings of male and female gender terms for configuration ``Gender 1 - Occupation 2'', split across stereotypically male and female jobs for the baseline and bias-mitigated Llama2 models.}
    \label{fig:llama2-cosine-dist-g1-o2}
\end{figure}

Figures \ref{fig:llama2-cosine-dist-g2-o2} and \ref{fig:llama2-cosine-dist-g1-o2} illustrate the distributions of cosine similarity values between embeddings of female and male gendered terms and occupation terms, split across stereotypically male and female occupations, for two configurations: ``Gender 1 – Occupation 2'' (where the gender term appears in the first sentence and the occupation in the second sentence) and ``Gender 2 – Occupation 2'' (where both terms appear in the second sentence). These configurations are selected to capture mutual semantic influence between gender and occupation terms as described in Section \ref{sec:models-and-data}. The analysis of cosine similarity distributions is complemented by the KS test statistics (results shown in Tables \ref{tab:ks-llama-g2o2} and \ref{tab:ks-llama-g1o2}), which provide statistical evidence of shifts in embedding space before and after applying bias mitigation to the Llama2 models. 


In the case of stereotypically female occupations from configuration ``Gender 2 – Occupation 2'' (Figure \ref{fig:llama2-cosine-dist-g2-o2}), the baseline Llama2 model (bottom left plot) reveals disparity in the cosine similarity distributions between male and female gender terms. Specifically, female terms exhibit stronger alignment with stereotypically female occupations, reflecting a gendered semantic association embedded in the model. Following bias mitigation (bottom right plot), these distributions become notably more aligned, indicating a reduction in the model's gender-specific associations. Additionally, the mitigation reduces the multi-modal nature of the distributions, suggesting that gender terms no longer cluster into subgroups that are more or less associated with stereotypical roles. This convergence points to a more uniform and balanced treatment of gender terms in relation to occupations. 

These observations are quantitatively supported by KS test results (Table \ref{tab:ks-llama-g2o2}). The baseline model shows a statistically significant difference between male and female terms (D = 0.2580, p < 0.0001), confirming the disparity observed in the KDE plot. After bias mitigation, this difference is reduced but remains statistically significant (D = 0.1510, p < 0.0001), reflecting improved, but not entirely neutral, alignment. Moreover, significant differences are observed when comparing distributions of baseline and mitigated embeddings for both female (D = 0.3340, p < 0.0001) and male terms (D = 0.4310, p < 0.0001), indicating that the mitigation strategy alters the embedding space for both genders.

A similar pattern emerges for stereotypically male occupations. In the baseline model (top left plot), a bi-modal distribution appears, where female terms unexpectedly show greater association with male jobs, and both gender groups contain subclusters of terms with differing degrees of occupational association. Post-mitigation (top right plot), the distributions converge towards unimodality with reduced disparity across and within gender groups, suggesting more balanced semantic representations. This is also supported by the KS test results: male vs. female distributions are significantly different in both the baseline (D = 0.2230, p < 0.0001) and mitigated models (D = 0.1670, p < 0.0001), but the reduction in D-statistic indicates improved alignment. Likewise, significant changes in distributions for female (D = 0.3150, p < 0.0001) and male terms (D = 0.3570, p < 0.0001) across both models confirm the effectiveness of mitigation.

\begin{table}[!tb]
    \scriptsize
    \centering
    \begin{tabular}{p{2.5cm}lrr}
        \toprule
        \textbf{Job Type} & \textbf{Comparison} & \textbf{D-statistic} & \textbf{p-value} \\
        \midrule
        \multirow{4}{*}{\shortstack{Stereotypically\\Male Jobs}} 
            & Female Terms vs Male Terms (Baseline)       & 0.2230 & $<$0.0001 \\
            & Female Terms vs Male Terms (Mitigated)      & 0.1670 & $<$0.0001 \\
            & Female Terms: Baseline vs Mitigated   & 0.3150 & $<$0.0001 \\
            & Male Terms: Baseline vs Mitigated     & 0.3570 & $<$0.0001 \\
        \midrule
        \multirow{4}{*}{\shortstack{Stereotypically\\Female Jobs}}
            & Female Terms vs Male Terms (Baseline)       & 0.2580 & $<$0.0001 \\
            & Female Terms vs Male Terms (Mitigated)      & 0.1510 & $<$0.0001 \\
            & Female Terms: Baseline vs Mitigated   & 0.3340 & $<$0.0001 \\
            & Male Terms: Baseline vs Mitigated     & 0.4310 & $<$0.0001 \\
        \bottomrule
    \end{tabular}
    \caption{Kolmogorov--Smirnov test results for cosine similarity distributions for configuration ``Gender 2 - Occupation 2'' with Llama2 models.}
    \label{tab:ks-llama-g2o2}
\end{table}

In the ``Gender 1 – Occupation 2'' configuration (Figure \ref{fig:llama2-cosine-dist-g1-o2} visualizes the cosine similarity distributions across stereotypical occupations while Table \ref{tab:ks-llama-g1o2} presents the KS test results), the distributions of cosine similarities between gender terms and both stereotypically male- and female-associated occupations appear less biased compared to the ``Gender 2 – Occupation 2'' scenario. Prior to bias mitigation, the similarity scores exhibit broader and more dispersed distributions, suggesting higher variability and weaker gender–occupation alignment. The baseline model shows a small but statistically significant difference between male and female term distributions for stereotypically male occupations (D = 0.0920, p = 0.0004), while the difference is not statistically significant for stereotypically female occupations (D = 0.0490, p = 0.1812). This indicates that although some gendered associations exist, they are less pronounced in this configuration.

Following bias mitigation, the similarity distributions become more concentrated around the mean and exhibit reduced disparity between male and female terms across both job categories, visually indicating a fairer embedding space. This is supported by the KS results: for stereotypically male occupations, the difference between male and female terms remains statistically significant but is slightly reduced (D = 0.0780, p = 0.0045). For stereotypically female occupations, the difference becomes statistically significant post-mitigation (D = 0.0840, p = 0.0017), although the effect size remains small. Additionally, significant shifts are observed between baseline and mitigated distributions for both female terms (D = 0.1770 for male-associated jobs; D = 0.1920 for female-associated jobs; p < 0.0001) and male terms (D = 0.1670 for male-associated jobs; D = 0.1130 for female-associated jobs; p < 0.0001), confirming that the mitigation technique introduces meaningful changes in the model’s representations.

Overall, these results demonstrate that while baseline decoder-only models such as Llama2 can exhibit a certain amount of gender bias in their association with occupational terms (based on configuration), these can be mitigated through targeted intervention.

\begin{table}[!tb]
    \scriptsize
    \centering
    \begin{tabular}{p{2.5cm}lrr}
        \toprule
        \textbf{Job Type} & \textbf{Comparison} & \textbf{D-statistic} & \textbf{p-value} \\
        \midrule
        \multirow{4}{*}{\shortstack{Stereotypically\\Male Jobs}}
            & Female Terms vs Male Terms (Baseline)       & 0.0920 & 0.0004\\
            & Female Terms vs Male Terms (Mitigated)      & 0.0780 & 0.0045\\
            & Female Terms: Baseline vs Mitigated         & 0.1770 & $<$0.0001\\
            & Male Terms: Baseline vs Mitigated           & 0.1670 & $<$0.0001\\
        \midrule
        \multirow{4}{*}{\shortstack{Stereotypically\\Female Jobs}}
            & Female Terms vs Male Terms (Baseline)       & 0.0490 & 0.1812\\
            & Female Terms vs Male Terms (Mitigated)      & 0.0840 & 0.0017\\
            & Female Terms: Baseline vs Mitigated         & 0.1920 & $<$0.0001\\
            & Male Terms: Baseline vs Mitigated           & 0.1130 & $<$0.0001\\
        \bottomrule
    \end{tabular}
    \caption{Kolmogorov--Smirnov test results for cosine similarity distributions for configuration ``Gender 1 - Occupation 2'' with Llama2 models.}
    \label{tab:ks-llama-g1o2}
\end{table}

\section{Conclusion}
This work demonstrates that gender bias mitigation in foundation models can manifest in measurable shifts in the embedding space, specifically in the associations between embeddings of gender terms and stereotypically gendered occupations. By investigating the embedding space of both encoder- and decoder-only foundation models pre- and post-bias mitigation, we discovered that association disparities between gender and occupation terms were consistently reduced post-bias mitigation. The reduction in cosine similarity disparity indicates that the embedding spaces of the mitigated models encode less gender-specific bias, offering an explanation for successful gender bias mitigation. These results contribute to our understanding of how bias-mitigation interventions transform the internal representations of foundation models and provide an interpretable framework to assess the effectiveness of these interventions through embedding space analysis.

Despite the promising findings, certain limitations to this work suggest directions for future research. Firstly, the tested strategies were primarily evaluated on their ability to mitigate gender bias. Investigating how the embedding space reshapes itself for other dimensions (e.g., race, age) could provide a broader understanding of embedding space dynamics. Moreover, future studies should connect the shifts in the embedding space to downstream task performance to help establish whether bias-mitigation approaches introducing fairer shifts in the embedding space consistently result in fairer behaviour in practical applications. Finally, future studies will benefit from testing across a wide range of foundation model types (beyond single encoder- and decoder-only model pairs) to evaluate whether positive outcomes in the embedding space based on targeted bias-mitigation transfer to other architectures and models.
\begin{credits}
\subsubsection{\ackname} This work was funded by the European Union’s Horizon Europe research and innovation programme under grant agreement no. 101070568 (AutoFair) and Taighde Éireann – Research Ireland under Grant number 18/CRT/6183. For the purpose of Open Access, the author has applied a CC BY public copyright licence to any Author Accepted Manuscript version arising from this submission.

\subsubsection{\discintname} The authors have no competing interests to declare that are relevant to the content of this article.
\end{credits}

%
%
%
\bibliographystyle{splncs04}
\bibliography{mybibliography}

\end{document}